\documentclass{article}
\usepackage{acra}
\usepackage{amsmath,amssymb}
\usepackage{graphicx}
\usepackage{booktabs}
\usepackage{placeins}
\usepackage{balance}
\usepackage[hidelinks]{hyperref}

\usepackage{etoolbox}

\makeatletter
\pretocmd{\@maketitle}{%
  \noindent
  {\small\bfseries
  Cite as: Huiting Xi, Marsha Mariya Kappan,
  Eduardo Benitez Sandoval, and Francisco Cruz.
  CJRR-Net: A Candidate-Joint Reprojection Ranking
  for Future-View Selection in Human Pose Estimation.
  \textit{Proceedings of the Australasian Conference
  on Robotics and Automation (ACRA)}, 2026.
  \par}
    \vspace{0.6em}
  \hrule
  \vspace{-2.6em}
}{}{}
\makeatother

\title{CJRR-Net: A Candidate-Joint Reprojection Ranking for Future-View Selection in Human Pose Estimation}
\author{
Huiting Xi$^{1}$,
Marsha Mariya Kappan$^{1}$,
Eduardo Benitez Sandoval$^{2}$,
Francisco Cruz$^{1,3}$ \\
$^{1}$School of Computer Science and Engineering,
UNSW Sydney, Australia \\
$^{2}$School of Art \& Design,
UNSW Sydney, Australia \\
$^{3}$Escuela de Ingeniería,
Universidad Central de Chile, Santiago, Chile \\
huiting.xi@student.unsw.edu.au,
\{m.marsha\_mariya\_kappan, e.sandoval, f.cruz\}@unsw.edu.au
}

\begin{document}

\maketitle

\begin{abstract}
Multi-camera systems support performance recording, sports and human-movement
analysis, where selecting
an informative viewpoint is important for reliable human pose estimation. We propose the Candidate-Joint Reprojection Ranking Network (CJRR-Net) for future-view selection among eight calibrated cameras in MPI-INF-3DHP. The model combines current-view pose observations, camera configuration, and candidate-specific geometric features obtained by reprojecting the estimated 3D pose and its uncertainty into each candidate view. CJRR-Net achieves a selected-view 2D mean per-joint position error
(MPJPE) of $105.18$ pixels,
compared with $114.64$ pixels for the shared-input MLP and
$117.56$ pixels for adapted PoseErrNet.
It also achieves the lowest mean error among the listed policies in each evaluated occlusion group. A blender rollout illustrates repeated camera selection.
These findings support candidate-specific geometry for single-view
selection in the evaluated fixed eight-camera setting.
\end{abstract}

\section{Introduction}
Markerless human pose estimation provides an unobtrusive way to analyse human motion from RGB images without attaching reflective markers or inertial sensors to the subject~\cite{mehta2017mpi}. The reliability of these estimates depends in part on the visual evidence available for individual body joints~\cite{visibility-aware,tokenpose}. Changes in body orientation and articulation can create self-occlusion and projected joint overlap, while scene objects can hide limbs and image boundaries can truncate the body. A fixed viewpoint may therefore provide insufficient evidence for some joints~\cite{visibility-aware}, and pose-estimation accuracy can vary with the camera viewpoint~\cite{activemocap}.


We study future-view selection in a calibrated camera array, where each decision uses one current-view RGB image and selects one camera for the next observation. This setting is motivated by applications with a single-stream
processing budget or one selected output, such as robotic observation~\cite{activemocap,poseerrnet},
automatic camera selection in performance recording~\cite{stoll2023theater},
and motion tracking and kinematic measurements in
sports~\cite{pagnon2021pose2sim}. In such settings, processing all camera streams simultaneously may be
undesirable because of computation, bandwidth, or real-time constraints,
while the system ultimately requires only one view for the next observation
or output. Multi-view fusion can combine complementary evidence when multiple
images are available; our task instead selects a future view before acquiring its image.

Existing active perception methods for pose-driven camera selection use
uncertainty-aware viewpoint planning~\cite{activemocap} or predict
viewpoint-dependent pose-estimation error~\cite{poseerrnet}. Both approaches
must assess candidate views before their images are available. An estimated
3D pose provides a spatial hypothesis for this purpose, but its projected
joint locations, spatial extent, and potential self-occlusion vary across
candidate cameras. This motivates representing the estimated pose and its
uncertainty explicitly relative to each candidate camera.

\begin{figure}[t]
    \centering
    \includegraphics[width=\linewidth]{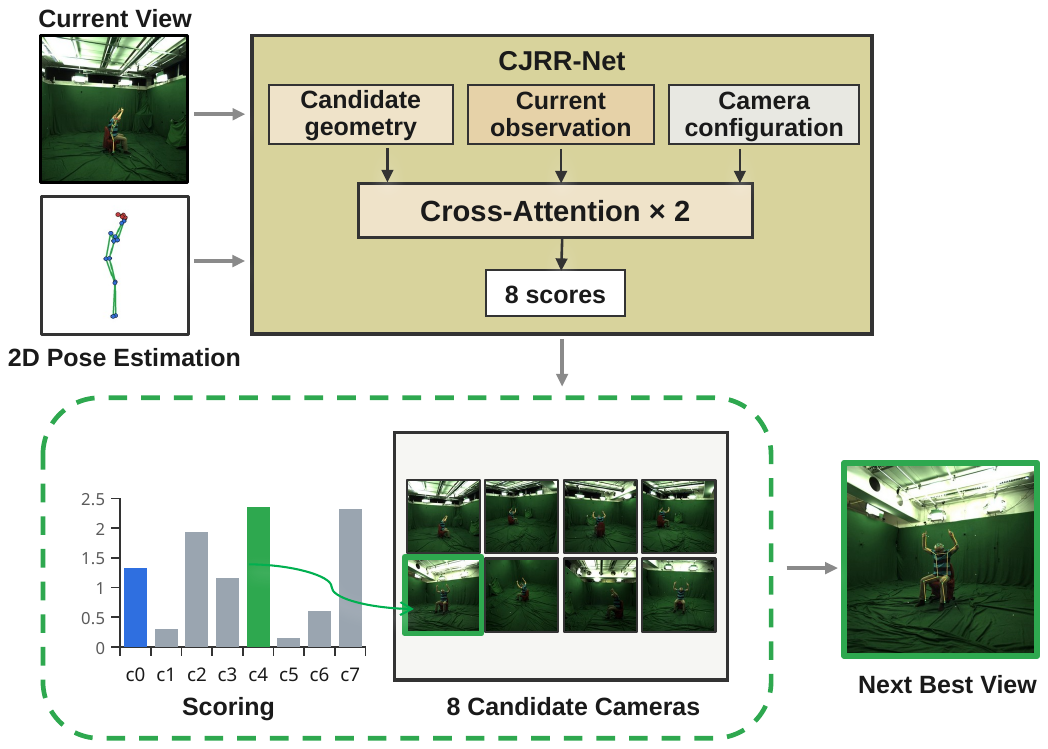}
    \caption{Overview of CJRR-Net for pose-driven camera selection. The current observation at time $t$ supplies the pose estimate used to construct geometric features for eight calibrated candidate cameras. These features are combined with observation context and camera configuration to select a future view for $t+\Delta$, with $\Delta=25$ frames. The score bars are schematic. Candidate images illustrate the camera set and are unavailable to the selector at
    decision time.}
    \label{fig:overview}
\end{figure}

In this work, we propose the candidate-joint Reprojection Ranking Network (CJRR-Net), a three-branch network, as illustrated in \hyperref[fig:overview]{Figure~\ref*{fig:overview}}, that combines current-view pose observations, candidate-camera configuration, and
candidate-specific geometric features to rank future viewpoints. The key design is an explicit geometric representation obtained by reprojecting the estimated 3D pose and its uncertainty into each calibrated candidate camera. It captures projected joint locations, visibility cues, spatial extent, and uncertainty. CJRR-Net combines these features with observation context and camera configuration to select a future viewpoint from the current observation.

Our contributions are threefold:
\begin{itemize}
\item A candidate-specific geometric representation that transforms an estimated 3D pose and its uncertainty into calibrated joint-level evidence for unobserved camera views.
\item A three-branch ranking framework that integrates this representation with current observations and camera configuration to select a future view without acquiring candidate images at decision time.
\item Shared-input comparisons, representation and branch controls, and occlusion-stratified evaluation that assess candidate geometry, together
with a blender rollout illustrating repeated camera selection.
\end{itemize}

\section{Related Work}
\label{sec:related}
\subsection{Human Pose Estimation under Occlusion}

Human pose estimation under occlusion requires reasoning about joints with unequal visual support. TokenPose~\cite{tokenpose} represents individual keypoints as tokens to learn visual cues and anatomical relationships. Visibility-aware reasoning~\cite{visibility-aware} predicts keypoint visibility to mask unreliable features from occluded joints and infer their locations from visible-joint context. These methods improve estimation from the available observation. For active perception, they motivate considering how a viewpoint change affects individual joints: a view that reveals one limb may obscure another, so aggregate visibility can
conceal differences relevant to camera selection.

Three-dimensional pose provides an intermediate representation for reasoning across viewpoints. Martinez et al.~\cite{martinez2017simple} lift 2D keypoints into 3D using a residual network. Neural Localizer Fields~\cite{sarandi2024nlf} localise arbitrary body points in 3D from RGB and predict pointwise uncertainty. MHDPose~\cite{kappan2026mhdpose} generates multiple plausible 3D
pose hypotheses from a 2D pose. These 3D pose representations support geometric reasoning across camera viewpoints and reduce the ambiguity of purely 2D observations.

However, these methods still rely on information available from the current observation. Under severe occlusion, some body parts may be invisible and their keypoint estimates unreliable~\cite{cheng2019occlusionaware}, making recovery more
dependent on learned priors than direct visual evidence. This motivates active viewpoint selection to acquire a more informative observation.

\subsection{Next-Best-View Selection for Human Pose Estimation}
Active viewpoint selection addresses occlusion by moving the camera or agent to acquire a more informative observation, rather than relying only on the pose estimator. Analytic criteria select viewpoints by optimising an explicit utility function. Information-driven criteria seek to reduce estimation uncertainty through entropy minimisation or expected information
gain~\cite{entoryminimum}, while related approaches use covariance-based optimality criteria~\cite{informationgain}. A representative example is ActiveMoCap~\cite{activemocap}, which approximates pose uncertainty using the Hessian of a reconstruction energy and plans future viewpoints to reduce 3D estimation uncertainty. For a perceptual view-preference objective, Kwon et
al.~\cite{kwon2022optimal} optimise camera position using projected
limb length, 2D person extent, and visible 3D bounding-box area. Together, these analytic methods show that candidate viewpoints can be
evaluated from pose uncertainty or projected geometry before acquisition.

Learned next-best-view (NBV) methods either predict the next viewpoint directly or estimate task-specific errors across viewpoints. For active 3D reconstruction,
NBVNet~\cite{nbvnet} uses a 3D convolutional network to predict the next
best view from a voxelised scene representation. For human pose estimation,
ACTOR and Pose-DRL use deep reinforcement learning to select informative
viewpoints for active triangulation or an underlying monocular pose
estimator~\cite{pirinen2019domes,PoseDRL}. More recently,
PoseErrNet~\cite{poseerrnet} maps normalised current 2D keypoints to a
viewpoint-dependent 2D pose-error field, using NeRF-based view synthesis for
supervision and a planner for camera motion.

These studies demonstrate that future viewpoints can be selected without first acquiring the corresponding images. However, they do not investigate whether reprojecting an estimated 3D pose and its uncertainty into each candidate camera can provide joint-level geometric evidence for more accurate view selection in a fixed, calibrated multi-camera system. Consequently,
they do not evaluate whether structured candidate-specific joint evidence
improves future-view ranking. We therefore introduce CJRR-Net to encode and aggregate this candidate-specific
joint evidence for future-view ranking.



\section{Method}
\begin{figure*}[t]
    \centering
    \includegraphics[width=\textwidth]{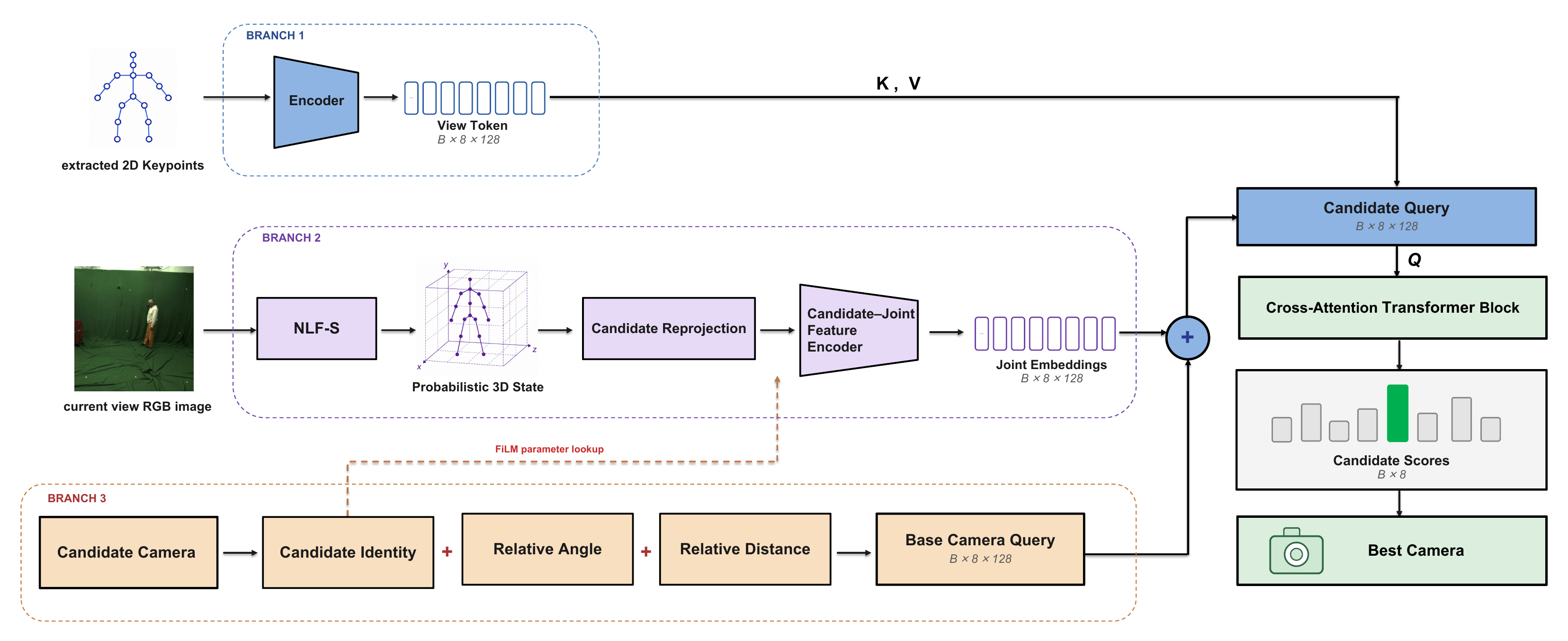}
    \caption{Architecture of CJRR-Net. View-context tokens, candidate-joint reprojection features, and camera-geometry queries from the three branches are fused through cross-attention to rank the candidate cameras.}
    \label{fig:architecture}
\end{figure*}

\hyperref[fig:architecture]{Figure~\ref*{fig:architecture}} presents the overall architecture of CJRR-Net. CJRR-Net combines current-view pose tokens, candidate-camera queries, and candidate-joint reprojection features through two cross-attention blocks. A shared head scores the candidates for future-view selection.

\subsection{Candidate-Joint Reprojection Field}
\label{sec:cjfield}

Using the frozen small variant of Neural Localizer Fields
(NLF-S)~\cite{sarandi2024nlf} and a separately trained RootHead, we estimate a probabilistic 3D pose and reproject its joint positions and uncertainties
into each candidate camera. Following the encode-then-aggregate principle for set-structured inputs~\cite{zaheer2017deepsets}, a shared MLP encodes
each candidate--joint feature, and learned attention
pooling~\cite{ilse2018attention} aggregates the resulting embeddings into
one representation per candidate camera.

\noindent\textbf{Probabilistic 3D State.}
Let $\mathbf I_t^r$ denote the current RGB observation captured by camera $r$ at time $t$. The frozen NLF-S estimator predicts camera-frame 3D joint coordinates, $\mathbf P^{\mathrm{NLF}}$, and joint-wise scalar uncertainties, $\boldsymbol\tau^{\mathrm{NLF}}$, as follows:
\begin{equation}
\left(
\mathbf P^{\mathrm{NLF}},
\boldsymbol\tau^{\mathrm{NLF}}
\right)
=
f_{\mathrm{NLF}}\!\left(\mathbf I_t^r\right).
\label{eq:nlf_state}
\end{equation}
Although NLF-S provides the articulated 3D pose, its global root translation can contain depth error. We therefore use a separately trained RootHead to refine the global root position. Let $\mathbf r^{\mathrm{NLF}}$ denote the original NLF-S root position, defined as the midpoint of the two hip joints, and $\widehat{\mathbf r}$ the corrected root position. For each joint $j$, $\boldsymbol{\mu}_j$ denotes the corrected 3D joint position, while $\boldsymbol{\Sigma}_j$ is its diagonal
covariance and represents the associated positional uncertainty. The resulting probabilistic 3D state is:

\begin{equation}
\begin{split}
\widehat{\mathbf r}
&=
\exp(\delta)\mathbf r^{\mathrm{NLF}}
+
[l_x,l_y,0]^{\top},\\
\boldsymbol\mu_j
&=
\mathbf P_j^{\mathrm{NLF}}
-\mathbf r^{\mathrm{NLF}}
+\widehat{\mathbf r},\\
\boldsymbol\Sigma_j
&=
\operatorname{diag}\!\left(
\boldsymbol\sigma_{\mathrm{root}}^2
+
(\tau_j^{\mathrm{NLF}}\mathbf s_{\mathrm{art}})^{\odot2}
\right).
\end{split}
\label{eq:probabilistic_state}
\end{equation}

Here, $\delta$ and $(l_x,l_y)$ denote the predicted log-depth scale correction and lateral corrections, respectively.
$\mathbf P_j^{\mathrm{NLF}}$ is the original NLF-S joint prediction, and $\boldsymbol\sigma_{\mathrm{root}}^2\in\mathbb R^3$ contains the predicted per-axis root variances.
The scalar NLF-S uncertainty $\tau_j^{\mathrm{NLF}}$ is mapped to
per-axis articulation standard deviations using the training-calibrated scale $\mathbf s_{\mathrm{art}}\in\mathbb R^3$.
The joint covariance combines root and articulation variances under a diagonal approximation that neglects their cross-covariances.
In this formulation, $(\cdot)^{\odot 2}$ denotes element-wise squaring, and $\operatorname{diag}(\cdot)$ forms a diagonal matrix with the input vector as its diagonal.


\noindent\textbf{Reprojection into Candidate Cameras.}

Let $(\mathbf R_v,\mathbf T_v)$ denote the extrinsics mapping world coordinates into camera $v$, and let $r$ and $c$ index the current and candidate cameras, respectively. Using the known calibration, the corrected joint mean and covariance are transformed from the current camera frame $r$ to the candidate camera frame $c$, yielding $\boldsymbol{\mu}_{c,j}$ and $\boldsymbol{\Sigma}_{c,j}$ expressed in the candidate camera's 3D coordinate system:
\begin{equation}
\begin{split}
\boldsymbol{\mu}_{c,j}
&=
\mathbf R_c\mathbf R_r^{\top}
(\boldsymbol{\mu}_j-\mathbf T_r)+\mathbf T_c,\\
\boldsymbol{\Sigma}_{c,j}
&=
\mathbf R_c\mathbf R_r^{\top}\boldsymbol{\Sigma}_j
\mathbf R_r\mathbf R_c^{\top}.
\end{split}
\label{eq:candidate_transform}
\end{equation}

The candidate-frame 3D mean $\boldsymbol\mu_{c,j}$ is projected to the 2D
image plane by $\pi_c$. The covariance is propagated through the
projection Jacobian to obtain image-space standard deviations:
\begin{equation}
\begin{split}
\begin{bmatrix}
u_{c,j} & v_{c,j}
\end{bmatrix}^{\top}
&=
\pi_c(\boldsymbol{\mu}_{c,j}),\\
\begin{bmatrix}
\sigma_{u,c,j} & \sigma_{v,c,j}
\end{bmatrix}^{\top}
&=
\sqrt{
\operatorname{diag}\!\left(
\mathbf J_{\pi_c}\boldsymbol{\Sigma}_{c,j}
\mathbf J_{\pi_c}^{\top}
\right)}.
\end{split}
\label{eq:candidate_projection}
\end{equation}


Reprojecting the estimated pose and uncertainty into each candidate camera gives the following candidate-joint feature vector:
\begin{equation}
\begin{split}
\mathbf F_{c,j}=[\;&
u_{c,j},v_{c,j},Z_{c,j},
\sigma_{u,c,j},\sigma_{v,c,j},\\
&p_{\mathrm{in},c,j},p_{\mathrm{self\text{-}occ},c,j},
d_{\mathrm{border},c,j},\\
&s_{\mathrm{person},c},s_{\mathrm{bone},c,j},
o_{\mathrm{overlap},c,j},q_j^{\mathrm{cur}}\;].
\end{split}
\label{eq:candidate_joint_feature}
\end{equation}

Here, $p_{\mathrm{in},c,j}$ is the product of the horizontal and vertical marginal Gaussian probabilities over $[0,2048]$ pixels, using the projected means and standard deviations. The border cue $d_{\mathrm{border},c,j}$ is
$\min(u_{c,j},2048-u_{c,j},v_{c,j},2048-v_{c,j})$, positive inside the image and negative outside. The current-view confidence $q_j^{\mathrm{cur}}$ is shared
across candidate cameras.

For the scale cues, $s_{\mathrm{person},c}$ measures the overall projected body extent using the bounding-box diagonal of all reprojected joints, including those outside the image. Let $w_c=\max_j u_{c,j}-\min_j u_{c,j}$ and $h_c=\max_j v_{c,j}-\min_j v_{c,j}$ denote the box width and height. The bone scale, $s_{\mathrm{bone},c,j}$,  measures local skeletal extent by averaging the projected lengths of edges incident
to joint $j$:
\[
\begin{aligned}
s_{\mathrm{person},c}
&=\sqrt{w_c^2+h_c^2+\epsilon},\\
s_{\mathrm{bone},c,j}
&=\frac{1}{|\mathcal N(j)|}
  \sum_{k\in\mathcal N(j)}\ell_{c,jk},
\end{aligned}
\]
where $\mathcal N(j)$ contains the direct COCO-17 skeleton
neighbours of joint $j$. The length $\ell_{c,jk}$ is computed as the square root of the squared image-plane distance between joints $j$ and $k$ plus $\epsilon=10^{-6}$ pixels squared. Both scales are expressed in pixels; person scale is shared across joints within each candidate, whereas bone scale is joint-specific.

For the self-occlusion cue, $p_{\mathrm{self\text{-}occ},c,j}$, and overlap cue, $o_{\mathrm{overlap},c,j}$, let $\mathcal A(j)$ contain all joints except joint $j$ and its direct skeleton neighbours.

We compute
\[
\begin{aligned}
p_{\mathrm{self\text{-}occ},c,j}
&=1-\prod_{k\in\mathcal A(j)}[1-Q(a_{c,jk}b_{c,jk})],\\
o_{\mathrm{overlap},c,j}
&=1-\prod_{k\in\mathcal A(j)}[1-Q(a_{c,jk})],
\end{aligned}
\]
where $\rho_{c,k}$ is obtained by multiplying an assumed
$80$ mm body radius by the horizontal focal length and dividing by the depth of joint $k$.
The proximity weight $a_{c,jk}$ is the exponential of the negative squared ratio of $\ell_{c,jk}$ to $\rho_{c,k}$.
The depth weight $b_{c,jk}$ applies a sigmoid to the depth of joint $j$ minus that of joint $k$, divided by $50$ mm.
Both scales are fixed heuristic parameters.
$Q$ clips its input to $[0,1-10^{-6}]$.
Overlap uses the same proximity weights without depth ordering.

\noindent\textbf{Joint-wise Encoding and Aggregation.}
\label{sec:joint_encoding}
For each candidate camera $c$, we form a separate set containing the feature vectors $\mathbf F_{c,j}$ of all body joints.
After obtaining $\mathbf F_{c,j}$, we first encode each joint feature separately and then aggregate the resulting joint representations to obtain a compact candidate-view representation. This preserves the feature combinations within each joint before aggregation.

Let $\widetilde{\mathbf F}_{c,j}=\mathcal N(\mathbf F_{c,j})$ denote fixed channel-wise normalisation. We first apply $\log(1+\max(x,0))$ to projected standard deviations and person/bone scales. All non-probability channels, including these transformed channels, are then centred by their training-set medians and divided by $\mathrm{IQR}/1.349$, where IQR is the
interquartile range; scales below $10^{-6}$ are replaced by $1$. The in-frame, self-occlusion, overlap, and confidence channels use $(x-0.5)/0.5$. All normalised values are clipped to $[-8,8]$ before FiLM.

To condition the normalised joint features on candidate-camera identity, we apply FiLM-style feature-wise affine modulation~\cite{perez2018film}. For each candidate camera $c$, a learned lookup table provides a scale vector $\boldsymbol\gamma_c$ and a shift vector $\boldsymbol\beta_c$,
which modulate the 12 feature channels before the shared joint MLP:
\begin{equation}
\mathbf a_{c,j}
=
(\mathbf 1+\boldsymbol\gamma_c)\odot
\widetilde{\mathbf F}_{c,j}
+\boldsymbol\beta_c.
\label{eq:input_film}
\end{equation}
Here $\boldsymbol\gamma_c,\boldsymbol\beta_c\in\mathbb R^{12}$ are lookup-table parameters indexed by candidate identity, shared across joints and initialised to zero. For eight cameras, the scale and shift lookup tables add $2\times8\times12=192$ parameters.

A shared MLP $f_{\mathrm{joint}}$ with widths
$12\rightarrow32\rightarrow32$, layer normalisation, and ReLU
maps each modulated feature $\mathbf a_{c,j}$ to a joint token
$\mathbf h_{c,j}$. Learned attention pooling then combines these
tokens into one representation $\mathbf z_c$ per candidate camera:
\begin{equation}
\begin{aligned}
\mathbf h_{c,j}&=f_{\mathrm{joint}}(\mathbf a_{c,j}),\\
\alpha_{c,j}&=\frac{\exp(g(\mathbf h_{c,j}))}
 {\sum_{k=1}^{J}\exp(g(\mathbf h_{c,k}))},\\
\mathbf z_c&=\sum_{j=1}^{J}\alpha_{c,j}\mathbf h_{c,j}
 \in\mathbb R^{32},
\end{aligned}
\label{eq:joint_pooling}
\end{equation}
where $J=17$ is the number of joints and $g$ is a shared linear scalar scorer. This construction retains per-joint feature combinations until after nonlinear encoding, allowing their contributions to be weighted during aggregation. It uses no explicit joint-identity embeddings or message passing between learned joint tokens, while skeleton relationships enter through feature construction. The pooled vector $\mathbf z_c$ is the output of the candidate-joint branch and is subsequently projected to enrich the query for camera $c$.

\subsection{Three-Branch Fusion and Candidate Ranking}
\label{sec:fusion}
The candidate-joint field described in \hyperref[sec:cjfield]{Section~\ref*{sec:cjfield}} is fused with camera information and the current-view observation to score each candidate.

\noindent\textbf{Current-Observation Branch.}
For each camera slot, the pose observation is represented by 17
two-dimensional keypoints and their associated confidence scores, forming
a tensor $\mathbf{x}_v\in\mathbb{R}^{17\times3}$. Unobserved camera slots
are represented by zero-filled inputs. A shared two-layer MLP encoder maps each flattened 51-dimensional input to a 128-dimensional view-context token. Stacking the tokens over all eight camera slots gives
$\mathbf T\in\mathbb R^{8\times128}$, which supplies the keys and
values for cross-attention.

\noindent\textbf{Candidate-Camera Branch.}
For current camera $r$ and candidate camera $c$, let
$\mathrm{az}_v$, $\mathrm{el}_v$, and $\mathbf{C}_v$ denote the azimuth,
elevation, and camera-centre position of camera $v$, respectively.
We define the normalised relative angular feature
$\Delta\mathbf{a}_{r,c}$ and normalised camera-centre distance
$d_{r,c}$ as
\begin{equation}
\begin{split}
\Delta\mathbf{a}_{r,c}
&=
\left[
\frac{\operatorname{wrap}(\mathrm{az}_c-\mathrm{az}_r)}{180^{\circ}},
\frac{\mathrm{el}_c-\mathrm{el}_r}{180^{\circ}}
\right]^{\top},\\
d_{r,c}
&=
\frac{\left\lVert\mathbf{C}_c-\mathbf{C}_r\right\rVert_2}{d_0}.
\end{split}
\label{eq:relative_camera_geometry}
\end{equation}
Here, angular differences are normalised by $180^\circ$, and
camera-centre coordinates are expressed in millimetres.
The fixed normalisation scale $d_0=8000\,\mathrm{mm}$
slightly exceeds the maximum pairwise camera-centre separation
of approximately $7.84$ m in the evaluated rig, yielding
dimensionless distances below $1$ for this configuration.

The corresponding camera query is
\begin{equation}
\mathbf q^{\mathrm{cam}}_{r,c}
=
\mathbf e_c
+
f_{\mathrm{geo}}(\Delta\mathbf a_{r,c})
+
f_{\mathrm{base}}(d_{r,c})
\in \mathbb R^{128},
\label{eq:camera_query}
\end{equation}
where $\mathbf e_c$ is a learned candidate-camera embedding, and
$f_{\mathrm{geo}}$ and $f_{\mathrm{base}}$ encode relative orientation
and camera-separation information, respectively.

\noindent\textbf{Fusion and View Scoring.}
The candidate-joint representation is projected and added to the camera
query:
\begin{equation}
\mathbf q^{\mathrm{enr}}_{r,c}
=\mathbf q^{\mathrm{cam}}_{r,c}
+f_{\mathrm{proj}}(\mathbf z_c)
\in\mathbb R^{128}.
\label{eq:enriched_query}
\end{equation}
The eight enriched queries are then processed by two cross-attention
blocks, using $\mathbf T$ as keys and values.  A shared scoring head maps each
fused query to a scalar logit $s_c$, and the selected next view is:
\begin{equation}
\hat{c}=\arg\max_{c\in\mathcal{C}} s_c.
\label{eq:view_selection}
\end{equation}

\section{Experiments}
\label{sec:experiments}
We evaluate CJRR-Net through quantitative comparisons and qualitative examples, focusing on the practical benefits of its candidate-joint representation.

\subsection{Experimental Setup}
\label{sec:exp-setup}
\noindent\textbf{Dataset and Sample Construction.}
We use both sequences (Seq1 and Seq2) from
MPI-INF-3DHP~\cite{mehta2017mpi}, with S1--S6 for training,
S7 for validation, and S8 for testing (4,254, 501, and 500
timestamps, respectively). Using eight synchronised, calibrated cameras $\mathcal C=\{0,1,2,4,5,6,7,8\}$, we pair observations at $t$ with candidate views at $t+25$.
Each test timestamp is evaluated from all eight current cameras, yielding 4,000 single-view decisions per run.
All candidates, including the current camera, remain eligible.

ViTPose~\cite{xu2022vitpose} extracts 17 current-view 2D keypoints and confidence scores; the same RGB image supplies the NLF-S/RootHead state
(Section~\ref{sec:cjfield}). Frozen NLF-S was pretrained on data including MPI-INF-3DHP~\cite{sarandi2024nlf}, but public documentation does not verify S8's exclusion.
Although S8 is held out from ranker training, possible pretraining overlap limits conclusions about generalisation to subjects unseen by the entire pipeline.
The shared-input MLP remains informative because it uses the same frozen front end.

Future-view supervision is the mean joint-wise Euclidean error between ViTPose detections and mapped 2D annotations in $2048\times2048$ pixel coordinates. The MPI-INF-3DHP head point is used for all five facial keypoints.
The minimum-error candidate defines the best-view label.
Future images and annotations are used only for supervision
and offline evaluation.

\noindent\textbf{Training Details.}
NLF-S and the separately trained RootHead remain fixed during ranker training. We use Adam with learning rate $10^{-3}$, weight decay $10^{-4}$, batch size 256, and dropout 0.1. Training runs for at most 100 epochs, while the learning rate is halved after five epochs without validation-loss improvement, and training stops after fifteen. The checkpoint with the lowest validation loss is retained. Training and validation loss use random-view dropout: the number of retained views is sampled uniformly from 1--8, and the lowest-index retained camera supplies the reference
state. Test inference always uses a single current view. Trained methods are evaluated with seeds $5,7,55,17,13,6,11,9,22,44$; results report the mean and sample standard deviation across these ten runs.

\noindent\textbf{Training Objective.}
The loss coefficients were established through staged development
experiments. The gap and regret weights were selected based on validation-set performance during model development, while the camera-pair weight was selected through a 15-seed sweep. The contributions of the regret and camera-pair terms were further assessed through an objective ablation. All coefficients were then held fixed across learned-method comparisons. The resulting objective combines
best-view classification, error-gap ranking, expected regret, and
camera-pair ranking:
\begin{equation}
\mathcal L=\mathcal L_{\mathrm{CE}}+0.1\mathcal L_{\mathrm{gap}}
          +0.5\mathcal L_{\mathrm{reg}}+0.4\mathcal L_{\mathrm{pair}}.
\label{eq:full_objective}
\end{equation}
Cross-entropy $\mathcal L_{\mathrm{CE}}$ targets the ground-truth best candidate, defined as the candidate with the lowest future pose error. The gap-weighted ranking term $\mathcal L_{\mathrm{gap}}$ penalises incorrect ordering between candidate views, while the expected-regret term $\mathcal L_{\mathrm{reg}}$ discourages selections with larger pose
errors. Because cameras 5 and 8 account for the largest shares of
ground-truth best-view labels under our dataset split and were frequently overselected during development, the pairwise ranking term $\mathcal L_{\mathrm{pair}}$ gives additional weight to comparisons in which the ground-truth best or second-best candidate has a lower error than either camera. The same objective is used for the representation controls and the shared-input MLP to ensure a consistent comparison.

\noindent\textbf{Evaluation Metrics.}
We evaluate camera-selection performance using predicted-view 2D mean per-joint position error (MPJPE), relative MPJPE reduction, Tol@15, and 0-Recall.

Predicted-view 2D MPJPE is the mean Euclidean distance between the predicted and ground-truth 2D joint locations in the selected future view, averaged over all joints and test decisions, where lower values indicate better selection.

\begin{table*}[!t]
  \centering
  \small
\caption{Overall future-view selection performance on the MPI-INF-3DHP
test set. Results for trained methods are reported as mean $\pm$
standard deviation over 10 random seeds. $\uparrow$ indicates that
higher values are better, and $\downarrow$ indicates that lower values
are better.}
    \label{tab:overall-performance}
  \begin{tabular}{lrrrr}
    \toprule
    Policy & \shortstack[r]{Predicted-view 2D\\MPJPE (px) $\downarrow$}
    & \shortstack[r]{Relative MPJPE\\reduction (\%) $\uparrow$}
    & Tol@15 (\%) $\uparrow$ & 0-Recall $\downarrow$ \\
    \midrule
Uniform random
    & $299.76\pm5.27$ & $0.00\pm1.76$ & $26.25\pm0.67$
    & $\mathbf{0.00\pm0.00}$ \\
    Geometric utility
    & $198.30\pm0.00$ & $33.85\pm0.00$ & $33.35\pm0.00$
    & $\mathbf{0.00\pm0.00}$ \\
ActiveMoCap-inspired planner & $181.29\pm13.55$ & $39.52\pm4.52$ & $40.37\pm4.02$ & $\mathbf{0.00\pm0.00}$ \\
PoseErrNet (adapted) & $117.56\pm5.02$ & $60.78\pm1.68$ & $53.81\pm0.45$ & $6.30\pm0.48$ \\
MLP with shared perception inputs & $114.64\pm3.42$ & $61.76\pm1.14$ & $55.15\pm0.80$ & $2.50\pm1.18$ \\
CJRR-Net (ours) & $\mathbf{105.18\pm2.61}$ & $\mathbf{64.91\pm0.87}$ & $\mathbf{58.17\pm0.59}$ & $\mathbf{0.00\pm0.00}$ \\

    \bottomrule
  \end{tabular}
\end{table*}

Relative MPJPE reduction is the percentage decrease in selected-view MPJPE
relative to uniform random selection. Tol@15 is the proportion of selections whose error is within 15 pixels of the minimum future-view error. 0-Recall counts candidate cameras with best-view-conditioned recall below $0.05\%$.

\subsection{Quantitative Results}
\noindent\textbf{Comparison with Baselines.}
As shown in \hyperref[tab:overall-performance]{Table~\ref*{tab:overall-performance}}, CJRR-Net is compared with uniform random selection, a geometric utility rule, an ActiveMoCap-inspired planner~\cite{activemocap}, and an adapted PoseErrNet~\cite{poseerrnet}. These baselines are evaluated using the same future-error targets while retaining their respective inputs and objectives. For a more direct representation comparison, we also include an MLP that uses the same 2D detections, confidence scores, NLF state, and camera information as CJRR-Net, and is trained with the same full objective.

CJRR-Net achieves the lowest mean error, the largest relative MPJPE reduction,
and the highest Tol@15 among the policies in the comparison. Its mean error is $10.54\%$ lower than adapted PoseErrNet and
$41.98\%$ lower than the ActiveMoCap-inspired planner.
Relative to the MLP sharing its perception inputs, CJRR-Net reduces error
by $9.46$ pixels and raises Tol@15 by $3.02$ percentage
points. This comparison supports
explicitly organising the available pose and camera information into candidate-specific geometric features.

Moreover, CJRR-Net maintains 0-Recall of zero in all ten runs, compared with $6.30\pm0.48$ near-zero-recall classes for adapted PoseErrNet and $2.50\pm1.18$ for the shared-input MLP. This indicates that the improvement is accompanied by broader camera-class coverage, rather than being achieved by concentrating predictions on a small subset of dominant candidate views.

\noindent\textbf{Contribution of Candidate-Specific Geometry.}
\label{sec:exp-candidate-geometry}
We examine both the contribution of the candidate geometry branch and how its information is represented, holding the NLF/RootHead state, detections, calibration, future-error targets, and full objective fixed.
\begin{table}[!htbp]
  \centering
  \small
  \setlength{\tabcolsep}{3pt}
  \caption{Candidate representation and branch controls.
  A: scalar summary; B: mean-12D; C: CJRR-Net.
  Results are mean $\pm$ sample standard deviation over ten seeds.
  Bold marks the best mean in each column.
  $\downarrow$: lower is better; $\uparrow$: higher is better.}
  \label{tab:v5-representation}
 \resizebox{\columnwidth}{!}{%
  \begin{tabular}{@{}lrrr@{}}
    \toprule
    Configuration
    & \shortstack{MPJPE\\(px) $\downarrow$}
    & \shortstack{Tol@15\\(\%) $\uparrow$}
    & \shortstack{0-Recall\\$\downarrow$} \\
    \midrule
    A: scalar summary
    & $117.58\pm1.46$
    & $54.25\pm0.63$
    & $2.10\pm1.20$ \\

    B: mean-12D
    & $105.19\pm2.52$
    & $57.66\pm0.63$
    & $0.50\pm0.71$ \\

    C: CJRR-Net
    & $105.18\pm2.61$
    & $\mathbf{58.17\pm0.59}$
    & $\mathbf{0.00\pm0.00}$ \\
    \midrule
    C w/o geometry
    & $116.05\pm1.57$
    & $54.52\pm0.59$
    & $2.40\pm0.70$ \\

    C w/o query
    & $108.88\pm2.66$
    & $56.23\pm0.70$
    & $0.80\pm0.63$ \\

    C w/o observation
    & $\mathbf{104.26\pm3.12}$
    & $57.63\pm0.54$
    & $0.70\pm0.48$ \\
    \bottomrule
  \end{tabular}
  }
\end{table}

\begin{figure}[!htbp]
  \centering
  \includegraphics[width=\linewidth]
    {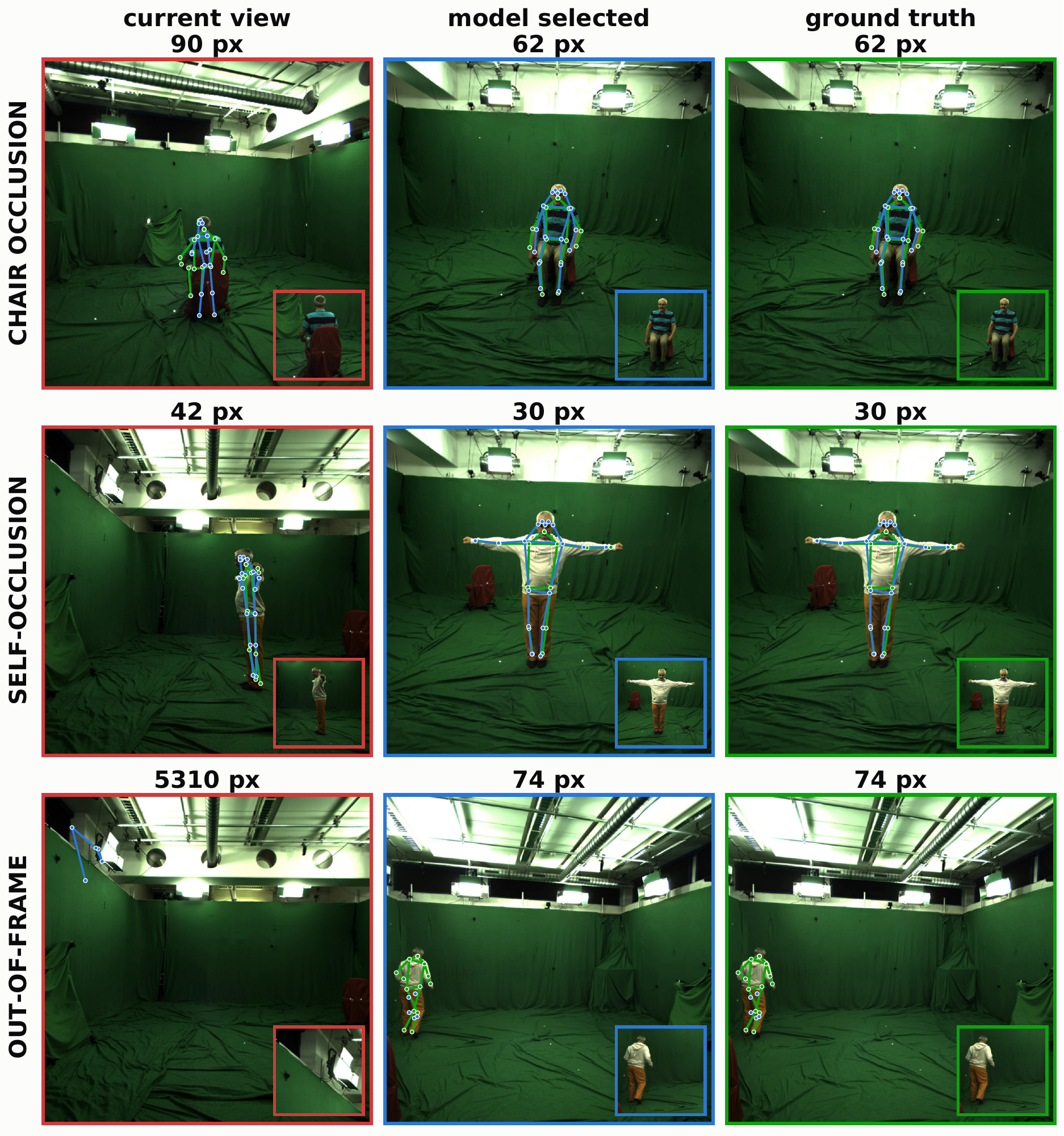}
  \caption{Successful examples of chair occlusion,
self-occlusion, and out-of-frame truncation (top to bottom).
Columns show the current, model-selected, and ground-truth views.
Values indicate 2D MPJPE in pixels.}
  \label{fig:v5-qualitative}
\end{figure}


Retaining all 12 geometric channels improves mean error over the scalar summary (Table~\ref{tab:v5-representation}). A uses a scalar confidence/visibility summary; B averages the normalised channels over joints before nonlinear encoding; C uses candidate-conditioned FiLM, joint-wise encoding, and attention pooling.

B and C nevertheless have nearly identical MPJPEs
($105.19\pm2.52$ versus $105.18\pm2.61$ px), compared with
$117.58\pm1.46$ px for A. These results support richer geometry but do not establish an MPJPE benefit from the additional architecture. C's mean Tol@15 is $0.51$ percentage points higher ($58.17\pm0.59\%$ versus $57.66\pm0.63\%$). This $0.51$-percentage-point gain is modest and should be
interpreted alongside variability across ten seeds.
However, C adds only 225 trainable parameters over B
(approximately $0.07\%$), including 192 for FiLM.
The higher observed Tol@15 therefore comes with minimal
additional parameter overhead, providing a practical reason to retain candidate-conditioned joint processing in the final model.

Removing the geometry or candidate-query branch raises mean MPJPE to $116.05$ and $108.88$ px, respectively.
Removing observation context instead lowers mean MPJPE to
$104.26\pm3.12$ px, but reduces mean Tol@15 to
$57.63\pm0.54\%$ and increases 0-Recall to $0.70\pm0.48$,
compared with $0.00\pm0.00$ for the full model.
The observation branch therefore shows a trade-off across mean
selection error, Tol@15, and camera-class coverage.

\begin{table*}[!htbp]
  \centering
  \small
  \caption{Inference-only candidate-availability analysis. Camera IDs are
original MPI-INF-3DHP IDs. Results are mean $\pm$ sample standard
deviation over ten checkpoints. Random MPJPE is computed within the same
permitted subset, and Tol@15 uses the minimum-error reference within that
subset; the model is not retrained.}
  \label{tab:candidate-availability}
  \begin{tabular}{clrrrr}
    \toprule
    Candidates & Camera IDs & \shortstack[r]{Random\\MPJPE (px)}
    & \shortstack[r]{Selected\\MPJPE (px) $\downarrow$}
    & \shortstack[r]{Relative MPJPE\\reduction (\%) $\uparrow$}
    & Tol@15 (\%) $\uparrow$ \\
    \midrule
    8 & $\{0,1,2,4,5,6,7,8\}$ & $299.76$
      & $105.18\pm2.61$ & $64.91$ & $58.17\pm0.59$ \\
    6 & $\{0,2,4,6,7,8\}$ & $338.14$
      & $132.83\pm2.57$ & $60.72$ & $60.17\pm0.52$ \\
    4 & $\{0,2,7,8\}$ & $345.49$
      & $134.25\pm2.65$ & $61.14$ & $62.30\pm0.34$ \\
    2 & $\{0,6\}$ & $258.31$
      & $158.06\pm3.48$ & $38.81$ & $85.64\pm1.44$ \\
    \bottomrule
  \end{tabular}
\end{table*}

\begin{figure*}[!htbp]
  \centering
  \includegraphics[width=\textwidth]{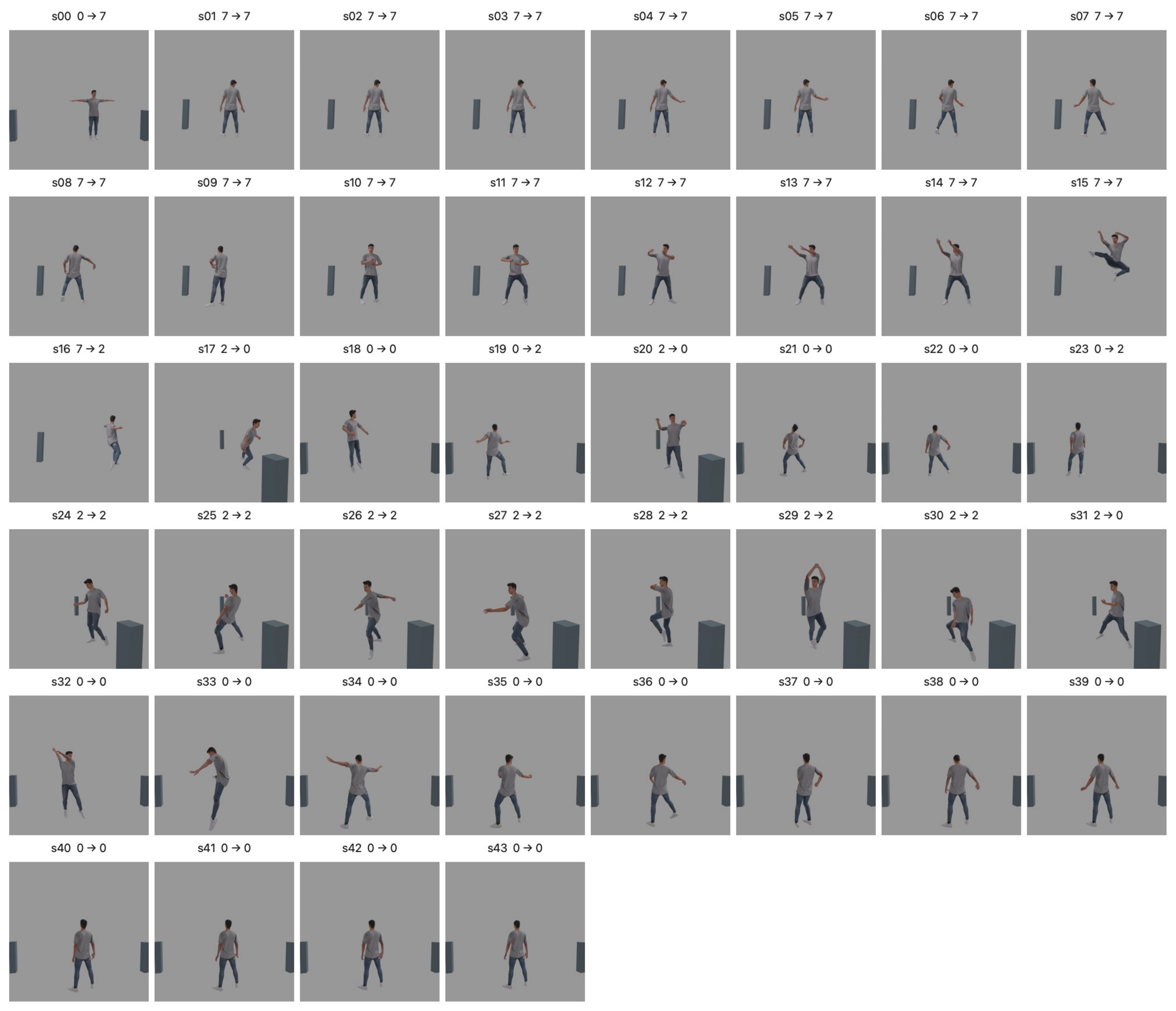}
  \caption{Complete 44-step closed-loop rollout in blender, ordered left-to-right and top-to-bottom. Each panel shows the current input image and labels the step and current-to-selected camera command. The selected camera becomes active 25 frames later (one second of simulated time). The 43 successive inputs follow the preceding camera commands. Visibility is illustrated qualitatively in this single rollout.}
  \label{fig:blender-closed-loop}
\end{figure*}

\begin{table*}[!t]
  \centering
  \small
  \caption{Selected-view 2D error (pixels) grouped by current-view occlusion. Trained methods report mean $\pm$ sample
  standard deviation over 10 seeds. The groups partition the same
  4,000 decisions as the overall comparison. This analysis covers the
  listed policies, their perception inputs and objectives follow
  \hyperref[tab:overall-performance]{Table~\ref*{tab:overall-performance}}.}
  \label{tab:v4-occlusion}
  \begin{tabular}{lrrr}
    \toprule
    Policy & Low (0--2) & Medium (3--5) & High ($\geq6$) \\
    \midrule
    Uniform random
    & $247.22\pm6.69$
    & $275.54\pm9.99$
    & $401.93\pm11.23$ \\
    
    Geometric utility
    & $162.45\pm0.00$
    & $193.74\pm0.00$
    & $253.63\pm0.00$ \\
    ActiveMoCap-inspired planner
    & $152.13\pm18.48$ & $163.85\pm12.90$ & $242.83\pm8.80$ \\
    PoseErrNet (adapted)
    & $97.87\pm2.89$ & $109.73\pm2.91$ & $154.37\pm11.65$ \\
    MLP with Shared Perception Inputs
    & $98.51\pm3.31$ & $102.82\pm2.18$ & $151.29\pm6.79$ \\
    CJRR-Net
    & $\mathbf{92.29\pm2.74}$
    & $\mathbf{99.24\pm2.16}$
    & $\mathbf{130.24\pm5.51}$ \\
    \bottomrule
  \end{tabular}
\end{table*}

\noindent\textbf{Performance under Occlusion.}
\label{sec:exp-occlusion}
To evaluate robustness under degraded current-view observations, we group test decisions by the number of joints occluded in the current view: Low (0--2), Medium (3--5), and High ($\geq6$). \hyperref[tab:v4-occlusion]{Table~\ref*{tab:v4-occlusion}} reports the performance of the evaluated methods within each occlusion level.

CJRR-Net achieves the lowest mean error in each occlusion group (Table~\ref{tab:v4-occlusion}). Its mean-error advantage over adapted PoseErrNet increases from $5.58$ pixels under low occlusion to $10.49$ and $24.13$ pixels under medium and high occlusion, respectively. This trend indicates that CJRR-Net retains a clear selection-quality advantage across occlusion levels, with the largest benefit observed under severe current-view occlusion.

\subsection{Reduced Candidate Availability}
\label{sec:candidate-availability}

The model computes all eight scores. Scores outside the permitted subset are set to $-\infty$ before selection, without retraining. This changes the selectable future views, whereas training-time view dropout changes the current observations. \hyperref[tab:candidate-availability]{Table~\ref*{tab:candidate-availability}} lists one fixed subset per size.


The eight-camera pool yields the lowest mean selected-view error among the tested subsets (Table~\ref{tab:candidate-availability}). Selection remains better than random with six, four, or two candidates, without retraining. These results demonstrate selection among subsets of known cameras, with one fixed layout evaluated per subset size.

\subsection{Qualitative Results}
\label{sec:exp-qualitative}

\noindent\textbf{View Selection under Visibility Degradation.}
\hyperref[fig:v5-qualitative]{Figure~\ref*{fig:v5-qualitative}} shows selected successful cases of external chair occlusion, self-occlusion, and out-of-frame truncation. Relative to the current views, the selected views provide
clearer lower-body visibility, arm-torso separation, and body framing,
respectively. These examples illustrate useful alternative observations
under the visibility degradations motivating active view selection.

\noindent\textbf{Closed-Loop Demonstration in Simulation.}
A closed-loop demonstration was implemented in blender 5.0.1 using the trained CJRR-Net pipeline, an animated human character from Renderpeople, and two fixed synthetic column-shaped occluders. We used the same eight calibrated
cameras as in the MPI evaluation. The character retained its FBX dance
animation, while its root followed an additional horizontal trajectory with a maximum displacement of $0.439$~m. The occluders were placed outside the character's motion envelope to produce viewpoint-dependent partial occlusion without intersecting the moving character.

\hyperref[fig:blender-closed-loop]{Figure~\ref*{fig:blender-closed-loop}} presents the complete decision sequence. Visual inspection shows that the subject remains visible throughout the displayed observations, with most views being either frontal or rear-facing and without complete occlusion by the columns. This example illustrates repeated perception and selection when each chosen camera supplies the next observation. The demonstration evaluates discrete camera switching in simulation.

\section{Conclusion and Future Work}
CJRR-Net selects future views by combining current-view pose observations, candidate-camera configuration, and geometric features obtained by reprojecting the estimated 3D pose and its uncertainty. Using a single current-view image and known camera calibration, it achieves a $64.91\%$ reduction in selected-view 2D MPJPE relative to uniform random selection on our eight-view MPI-INF-3DHP evaluation, outperforming the
listed baselines across the evaluated occlusion levels. Ablations support
the contribution of candidate geometry, and a blender rollout illustrates
repeated camera selection.

The evaluation is limited to one subject held out from ranker training and one fixed candidate set per camera count. Possible overlap between NLF-S pretraining and the S8 test subject limits conclusions about generalisation to subjects unseen by the entire pipeline. Closed-loop performance is demonstrated qualitatively. Future work will evaluate multiple layouts per camera count, retrain models for the corresponding candidate sets, and revisit camera-identity embeddings and candidate-specific FiLM for unseen
configurations. We will also measure closed-loop pose error and switching frequency against current-camera retention and fixed-camera baselines on additional subjects and real recordings.

With denser training coverage of camera positions and orientations, the geometric representation and prediction head could be adapted for
continuous viewpoint prediction. Moving-camera extensions should integrate temporal pose prediction and scene-occlusion reasoning with motion planning that accounts for travel time and collision constraints. Potential applications include gesture-aware robot interaction, pose-aware performance capture, and sports-motion analysis, with downstream benefits evaluated in each setting.



\begin{thebibliography}{}

\bibitem[\protect\citeauthoryear{Chen \bgroup \em et al.\egroup}{2024}]{poseerrnet}
Jingxi Chen, Botao He, Chahat~Deep Singh, Cornelia Ferm{\"u}ller, and
  Yiannis Aloimonos.
\newblock Active human pose estimation via an autonomous {UAV} agent.
\newblock In {\em Proceedings of the 2024 IEEE/RSJ International Conference
  on Intelligent Robots and Systems (IROS)}, pages 7801--7808, 2024.


\bibitem[\protect\citeauthoryear{Ilse \bgroup \em et al.\egroup}{2018}]{ilse2018attention}
Maximilian Ilse, Jakub~M. Tomczak, and Max Welling.
\newblock Attention-based deep multiple instance learning.
\newblock In {\em Proceedings of the 35th International Conference on
  Machine Learning}, volume 80 of {\em Proceedings of Machine Learning
  Research}, pages 2127--2136, 2018.

\bibitem[\protect\citeauthoryear{Kiciroglu \bgroup \em et al.\egroup}{2020}]{activemocap}
Sena Kiciroglu, Helge Rhodin, Sudipta~N. Sinha, Mathieu Salzmann, and
  Pascal Fua.
\newblock {ActiveMoCap}: Optimized viewpoint selection for active human
  motion capture.
\newblock In {\em Proceedings of the IEEE/CVF Conference on Computer Vision
  and Pattern Recognition (CVPR)}, pages 103--112, 2020.


\bibitem[\protect\citeauthoryear{Li \bgroup \em et al.\egroup}{2021}]{tokenpose}
Yanjie Li, Shoukui Zhang, Zhicheng Wang, Sen Yang, Wankou Yang,
  Shu-Tao Xia, and Erjin Zhou.
\newblock {TokenPose}: Learning keypoint tokens for human pose estimation.
\newblock In {\em Proceedings of the IEEE/CVF International Conference on
  Computer Vision (ICCV)}, pages 11313--11322, 2021.

\bibitem[\protect\citeauthoryear{Martinez \bgroup \em et al.\egroup}{2017}]{martinez2017simple}
Julieta Martinez, Rayat Hossain, Javier Romero, and James~J. Little.
\newblock A simple yet effective baseline for 3D human pose estimation.
\newblock In {\em Proceedings of the IEEE International Conference on
  Computer Vision (ICCV)}, pages 2640--2649, 2017.

\bibitem[\protect\citeauthoryear{Mehta \bgroup \em et al.\egroup}{2017}]{mehta2017mpi}
Dushyant Mehta, Helge Rhodin, Dan Casas, Pascal Fua, Oleksandr Sotnychenko,
  Weipeng Xu, and Christian Theobalt.
\newblock Monocular 3D human pose estimation in the wild using improved
  {CNN} supervision.
\newblock In {\em Proceedings of the International Conference on 3D Vision
  (3DV)}, pages 506--516, 2017.

\bibitem[\protect\citeauthoryear{Perez \bgroup \em et al.\egroup}{2018}]{perez2018film}
Ethan Perez, Florian Strub, Harm de~Vries, Vincent Dumoulin, and
  Aaron~C. Courville.
\newblock {FiLM}: Visual reasoning with a general conditioning layer.
\newblock In {\em Proceedings of the Thirty-Second AAAI Conference on
  Artificial Intelligence}, pages 3942--3951, 2018.

\bibitem[\protect\citeauthoryear{S{\'a}r{\'a}ndi and Pons-Moll}{2024}]{sarandi2024nlf}
Istv{\'a}n S{\'a}r{\'a}ndi and Gerard Pons-Moll.
\newblock Neural Localizer Fields for continuous 3D human pose and shape
  estimation.
\newblock In {\em Advances in Neural Information Processing Systems},
  volume 37, pages 140032--140065, 2024.

\bibitem[\protect\citeauthoryear{Sommerlade and Reid}{2008}]{entoryminimum}
Eric Sommerlade and Ian Reid.
\newblock Information-theoretic active scene exploration.
\newblock In {\em Proceedings of the IEEE Conference on Computer Vision and
  Pattern Recognition (CVPR)}, pages 1--7, 2008.

\bibitem[\protect\citeauthoryear{Sun \bgroup \em et al.\egroup}{2024}]{visibility-aware}
Pengzhan Sun, Kerui Gu, Yunsong Wang, Linlin Yang, and Angela Yao.
\newblock Rethinking visibility in human pose estimation: Occluded pose
  reasoning via transformers.
\newblock In {\em Proceedings of the IEEE/CVF Winter Conference on
  Applications of Computer Vision (WACV)}, pages 5903--5912, 2024.

\bibitem[\protect\citeauthoryear{Vasquez-Gomez \bgroup \em et al.\egroup}{2021}]{nbvnet}
Juan Irving Vasquez-Gomez, David Ernesto Troncoso Romero, Israel Becerra,
  Luis Enrique Sucar, and Rafael Murrieta-Cid.
\newblock Next-best-view regression using a 3D convolutional neural network.
\newblock {\em Machine Vision and Applications}, 32:42, 2021.

\bibitem[\protect\citeauthoryear{Wenhardt \bgroup \em et al.\egroup}{2007}]{informationgain}
Stefan Wenhardt, Benjamin Deutsch, Elli Angelopoulou, and Heinrich Niemann.
\newblock Active visual object reconstruction using D-, E-, and T-optimal
  next best views.
\newblock In {\em Proceedings of the IEEE Conference on Computer Vision and
  Pattern Recognition (CVPR)}, pages 1--7, 2007.

\bibitem[\protect\citeauthoryear{Xu \bgroup \em et al.\egroup}{2022}]{xu2022vitpose}
Yufei Xu, Jing Zhang, Qiming Zhang, and Dacheng Tao.
\newblock {ViTPose}: Simple vision transformer baselines for human pose
  estimation.
\newblock In {\em Advances in Neural Information Processing Systems},
  volume 35, pages 38571--38584, 2022.

\bibitem[\protect\citeauthoryear{Zaheer \bgroup \em et al.\egroup}{2017}]{zaheer2017deepsets}
Manzil Zaheer, Satwik Kottur, Siamak Ravanbakhsh, Barnabas Poczos,
  Ruslan Salakhutdinov, and Alexander~J. Smola.
\newblock Deep Sets.
\newblock In {\em Advances in Neural Information Processing Systems},
  volume 30, pages 3391--3401, 2017.

\bibitem[\protect\citeauthoryear{Pagnon \bgroup \em et al.\egroup
  }{2021}]{pagnon2021pose2sim}
David Pagnon, Mathieu Domalain, and Lionel Reveret.
\newblock {Pose2Sim}: An end-to-end workflow for 3D markerless sports
  kinematics---part 1: Robustness.
\newblock {\em Sensors}, 21(19):6530, 2021.

\bibitem[\protect\citeauthoryear{Stoll \bgroup \em et al.\egroup
  }{2023}]{stoll2023theater}
Eckhard Stoll, Stephan Breide, Steve G{\"o}ring, and Alexander Raake.
\newblock Automatic camera selection, shot size, and video editing in
  theater multi-camera recordings.
\newblock {\em IEEE Access}, 11:96673--96692, 2023.

\bibitem[\protect\citeauthoryear{Kappan \bgroup \em et al.\egroup
  }{2026}]{kappan2026mhdpose}
Marsha Mariya Kappan, Eduardo Benitez Sandoval, Erik Meijering, and
  Francisco Cruz.
\newblock {MHDPose}: Multi-hypothesis 3D human pose estimation using
  bidirectional Mamba diffusion models.
\newblock {\em Pattern Recognition}, 180:114396, 2026.

\bibitem[\protect\citeauthoryear{Kwon \bgroup \em et al.\egroup
  }{2022}]{kwon2022optimal}
Beom Kwon, Jungwoo Huh, Kyoungoh Lee, and Sanghoon Lee.
\newblock Optimal camera point selection toward the most preferable view of
  3-D human pose.
\newblock {\em IEEE Transactions on Systems, Man, and Cybernetics: Systems},
  52(1):533--553, 2022.

  \bibitem[\protect\citeauthoryear{Pirinen \bgroup \em et al.\egroup
  }{2019}]{pirinen2019domes}
Aleksis Pirinen, Erik G{\"a}rtner, and Cristian Sminchisescu.
\newblock Domes to drones: Self-supervised active triangulation for 3D human
  pose reconstruction.
\newblock In {\em Advances in Neural Information Processing Systems},
  volume 32, pages 3907--3917, 2019.

\bibitem[\protect\citeauthoryear{G{\"a}rtner \bgroup \em et al.\egroup
  }{2020}]{PoseDRL}
Erik G{\"a}rtner, Aleksis Pirinen, and Cristian Sminchisescu.
\newblock Deep reinforcement learning for active human pose estimation.
\newblock In {\em Proceedings of the AAAI Conference on Artificial
  Intelligence}, volume 34, number 7, pages 10835--10844, 2020.

\bibitem[\protect\citeauthoryear{Cheng \bgroup \em et al.\egroup
  }{2019}]{cheng2019occlusionaware}
Yu Cheng, Bo Yang, Bo Wang, Wending Yan, and Robby~T. Tan.
\newblock Occlusion-aware networks for 3D human pose estimation in video.
\newblock In {\em Proceedings of the IEEE/CVF International Conference on
  Computer Vision (ICCV)}, pages 723--732, 2019.

\end{thebibliography}
\balance

\end{document}